\def\year{2021}\relax
\documentclass[letterpaper]{article} % DO NOT CHANGE THIS
\usepackage{aaai21}  % DO NOT CHANGE THIS
\usepackage{times}  % DO NOT CHANGE THIS
\usepackage{helvet} % DO NOT CHANGE THIS
\usepackage{courier}  % DO NOT CHANGE THIS
\usepackage[hyphens]{url}  % DO NOT CHANGE THIS
\usepackage{graphicx} % DO NOT CHANGE THIS
\usepackage{natbib}  % DO NOT CHANGE THIS AND DO NOT ADD ANY OPTIONS TO IT
\usepackage{caption} % DO NOT CHANGE THIS AND DO NOT ADD ANY OPTIONS TO IT
\title{Requirements for Open Political Information: Transparency Beyond Open Data}
\author{
    Andong Luis Li Zhao \textsuperscript{\rm 1},
    Andrew Paley \textsuperscript{\rm 1}, 
    Rachel Adler \textsuperscript{\rm 1} \textsuperscript{\rm 2},
    Harper Pack \textsuperscript{\rm 1}, \\
    Sergio Servantez \textsuperscript{\rm 1},
    Alexander Einarsson \textsuperscript{\rm 1},
    Cameron Barrie \textsuperscript{\rm 1},
    Marko Sterbentz \textsuperscript{\rm 1}, \\
    Kristian Hammond \textsuperscript{\rm 1}
 \\
}
\affiliations{
    \textsuperscript{\rm 1}Northwestern University, 
    \textsuperscript{\rm 2}Northeastern Illinois University\\
    \{andong, andrewpaley\}@u.northwestern.edu, 
    r-adler@neiu.edu, 
    harper.pack@northwestern.edu, \\
    \{servantez, aeinarsson, cameron.barrie, marko.sterbentz\}@u.northwestern.edu, \\
    Kristian.Hammond@northwestern.edu

}
\begin{document}

\maketitle

\begin{abstract}

A politically informed citizenry is imperative for a well-developed democracy. While the US government has pursued policies for open data, these efforts have been insufficient in achieving an open government because only people with technical and domain knowledge can access information in the data. In this work, we conduct user interviews to identify wants and needs among stakeholders. We further use this information to sketch out the foundational requirements for a functional political information technical system.

\end{abstract}

\section{Introduction}
A modern democratic country’s political system is an intricate tapestry of multiple, often conflicting, tangible and intangible institutions, including an organized state, an independent bureaucracy, an adherence to rule of law \cite{fukuyama2014political}, and, ideally, an informed citizenry. 

To support such a citizenry, the UN Sustainable Development Goals make it clear: transparent government is a requirement of a well-developed and sustainable society. Specifically in the US, the principles of transparency have proliferated in various forms, from the Freedom of Information Act (FOIA) in 1966 \cite{yu2011new} to the more recent advent of open data initiatives \cite{Data.gov, fec.gov, mcdermott2010building}. There have also been significant third party efforts to make various forms of raw data available \cite{Historical, opensecrets, votesmart}. Across these initiatives, a common theme is a focus on issues of data access. 

Access to data is far from a solved issue -- sometimes the available data is constrained and requires specific requests \cite{kwoka2015foia}, sometimes it’s locked behind paywalls \cite{pacer}, and often available repositories are decentralized or in formats that are difficult to combine or use. Further, there are issues of scope: despite the above, federal data is significantly more available than data about state and local governments. 

Nevertheless, even if all of these problems are solved, ``open data'' alone is an insufficient policy for achieving ``open government” -- the actual requirement is \textit{open information}. This distinction is important; to a citizen without technical skills (or the resources to acquire them), a zip file of a hundred thousand documents or database credentials is a relatively opaque form of transparency. And, even for those with domain knowledge and technical skills -- such as data-driven journalists -- the sheer volume and pace of raw data generated can become insurmountable for one individual \cite{Historical}. In fact, it is people who can afford to hire teams of technical and domain experts who profit most from this opaque data  \cite{janssen2012benefits, kwoka2015foia}, indicating that open data alone might be reinforcing inequalities in the political system. 

Hence, truly open government requires that we rethink the means of access, moving from providing data to providing information. This is all the more important in our currently fraught information landscape. With the advent of social media misinformation and threats to basic institutions that we historically have taken for granted, societies around the world have realized the importance of upholding and strengthening our political institutions -- and part of doing so is making sure they’re accessible, understandable and transparent to all of their stakeholders. 

In this paper, we explore the issue of identifying the sociotechnical requirements for political information systems, taking into account new types of user experiences afforded by AI. As a first step, we focus our research on information about Illinois state politics. We conducted interviews with domain stakeholders, political scientists and journalists, to understand their needs. We report the current problems in the space of technical systems for political transparency based on real user feedback. Finally, we summarize our findings in a set of sociotechnical requirements that political information systems should fulfill to facilitate political transparency.

\section{Related Works}
The government itself keeps records available to all \cite{Data.gov, fec.gov}, although sometimes behind paywalls \cite{pacer}. This is an important first step. However, these have significant issues \cite{talukder2019determinants}, and require skills or resources to make use of the data \cite{janssen2012benefits, Kitchin2013}.

Outside of government, various civil institutions have initiatives aimed at access to campaign finance and legislative data at state and federal levels \cite{followthemoney, openstates, Store_2021}, supplemented by efforts to standardize data schemas \cite{opencivicdata}. However, technical skills and political knowledge are required to make use of these APIs. Further initiatives that allow users to track politicians and legislation \cite{Historical, opensecrets, votesmart} are mostly limited to federal officials and rely heavily on human-curated material, limiting scalability. Also, these initiatives appear to lack a user-centered design approach that leads to more usable interfaces \cite{norman1986user}.

In the AI and government space, there has been research exploring user needs \cite{smith2019designing}, with some of these works even proposing related solutions \cite{faruqe2020monitoring}. However, most of these works have not conducted user interviews and also tend to be broadly focused on public sector applications, yielding potentially overbroad requirements not informed by their stakeholders’ needs.

\section{User Interviews}

To ensure that we identify and address our users’ needs, we took a user-centered design approach. Our main goal for the interviews was to understand what questions users had regarding Illinois politics and the main impediments to getting answers . We also asked them to describe their background and skills to better understand their range of technical capabilities. These interviews lasted roughly 30 to 60 minutes.

\subsection{Stakeholders}
Our target users are people who have interest and expertise in political information but  lack either the technical skills necessary for more robust data wrangling and analysis  or the time or resources necessary to do so). We interviewed 15 users: 4 political science professors, 4 political science undergraduate students, 2 political science PhD students, 1 journalism professor, and 4 reporters. All but one user were based in Illinois (one political science professor was based in New York). We obtained this sample of users via our professional networks. Below are the highlights from the interviews categorized by the three main stakeholders.

\subsubsection{Journalists}

Journalists were one of our primary focuses because they exemplified our target user. All four journalists were focused on Illinois state or local politics, had interests in data-driven journalism, and often did their own entry-level analyses with platforms like Microsoft Excel. However, they all still had many questions that they were unable to answer due to lack of skills or resources to collect, collate, and analyze complex datasets.

\subsubsection{Professors}

Professors (5 total) were on the higher end of users who had technical and domain expertise. Three of them were interested in exploring questions about political trends in legislation, finances, and accountability to voters. Three professors discussed data challenges -- including staleness and lack of interoperability -- but only one professor had significant data analysis experience. 

The remaining two professors we interviewed had more methodological rather than domain-specific questions, including how to survey information from constituents, methods for modelling political behavior, and use of natural language processing to capture political information.

\subsubsection{Political Science Students}

Graduate students had some experience with research, technical tools, and the domain. They had systemic questions, and focused on political processes’ trends rather than election-focused or daily political questions. However, they still had significant issues with political transparency tools, especially due to their cumbersome user interfaces and the complexity of conducting database-wide and cross-dataset analyses.

Undergraduate students were the stakeholders who relied least on primary data for their political information, favoring lower-barrier sources, such as Twitter or other social media feeds. Almost all had minimal experience with data analysis, with only one student indicating they had performed statistical analysis to research a question.

\subsection{User Interview Themes}

\subsubsection{Questions they want to answer}

Of the users who were interested in Illinois State politics, the three most common topics of interest were the role money plays in politics, legislative process, and voting behavior. Questions include:
\begin{itemize}
    \item What groups of donors are most influential? How does this vary across time and political party?
    \item How does the donor profile of a legislator affect their voting pattern? Can we look at this relationship for different interest group and topics (e.g. environmental regulation, healthcare, immigration)?
    \item What bills are being repeatedly introduced to the legislature even though they consistently fail to pass? What bill changes were implemented such that these previously failing bills are able to pass?
    \item How do voting patterns change according to different demographics (e.g. race, rural vs. urban, economic class)?
\end{itemize}

There were also users who had interests in other domains such as international relations, landlord records, police misconduct, and criminal arrest records. 

\subsubsection{Working with primary data sources}

From our set of users, 6 of them had worked with or were familiar with political data sources such as the Illinois State Board of Elections (ISBE) or the Illinois General Assembly (ILGA). The other users dealt with more experimental datasets (e.g. research surveys) or obtained information from news sources (e.g. FiveThirtyEight or Twitter). 

The most common problems that these sources had were that the data was difficult to obtain and that it was difficult to understand the data and the platform. Obtaining the data from original sources is difficult for many reasons. For instance, one of the reporters mentioned wanting to work a story about pensions for public sector workers in Chicago, yet the data are scattered across many departments’ websites; similarly, we heard the Chicago police data was unlinked from the criminal court data of Cook County. The data can also be quite unreliable and out of date, such as that in the Chicago Data portal.

Perhaps the biggest challenge to overcome was that user experience was a significant barrier to entry. Though some of the websites, like the Center for Tax and Budget Accountability, had good capabilities and interfaces, many others had significant issues. In fact, one of the users said that the ILGA website was ``a nightmare” and that learning how to use the ISBE website took several years. 
 %Another user was even unable to get the data they wanted via a FOIA request. For instance, they wanted to ask about all limited liability companies (LLC) registered to a single address, but they were only able to request the address of a specific LLC. This would require them to have to request the address of all LLC’s, and then look at the subset of LLC’s that are registered to the address of interest.

\subsubsection{Working with auxiliary platforms}

The same set of 6 users who were familiar with primary political data sources were also familiar with auxiliary platforms that compile and process data from main sources; platforms like these include OpenStates, Reform for Illinois, and Legistar. 

Overall users reported more satisfaction with these platforms. However, some users were unsure if they could trust the platforms’ data. One user worked with the website FindMyLandlord, which helps users find Chicago building landlords, but they felt uncomfortable using it because it was unclear how the data was obtained and processed. Additionally, the user found errors in the data and were unable to suggest changes, further making them distrust the data and the platform. These issues sometimes were significant enough that users would rather deal with the problems of the primary sources to ensure reliability and trust in their data.

\subsubsection{Time, skills, and resource constraints}

Users reported that even after getting the needed data, they had difficulty using the data due to time, skills, and resource constraints. For many users, data analysis was not the primary function of their job nor was it part of their educational background. Hence, even entirely open, updated, and documented data would be difficult for them to take full advantage of, let alone stores of unstructured documents.

Understanding the data schema and properties was a significant issue for all users. Once the data was understood, users reported that they were able to run simple queries and  visualizations. The most advanced user was able to use Data Wrapper to display information; all other users did their analysis with platforms like Microsoft Excel. 

Some users reported they would like to have an experienced data scientist help them and some  went even further and wanted a tool that would just answer the questions they had, such as ``which donors gave money to legislator X and how did X vote on legislation related to the donors?”

\section{System Requirements}

From these user interviews, we identified main principles that political information systems should follow. These can be broken down into the four main areas of data requirements, usability, transparency, and domain personalization. Though our user interviews were mostly focused on Illinois state politics, these requirements broadly apply to political information systems, and we believe these requirements can be applied to other domains as well. However, consideration of other domains is outside of the scope of this research and will not be explored in this paper.

\subsubsection{Data Requirements}

\begin{enumerate}

    \item \textbf{Documentation}: Because political data are both timely and include various terms of art, having thorough documentation, such as data schemas and a glossary, helps avoid ambiguities and lowers the barrier to entry to new users.
    \item \textbf{Access to and interoperability with other data}: Since political data are scattered, it is critical that platforms can utilize decentralized data for more holistic analyses.This includes data on legislation, executive and judicial actions, electoral finances, electoral results, demographics, and media content. 
    \item \textbf{Access to up-to-date data}: Because political systems constantly generate new data, current approaches that focus on a data snapshot make it difficult to get information beyond that dataset’s scope.

\end{enumerate}

\subsubsection{Domain Personalization}

\begin{enumerate}
\setcounter{enumi}{3}
    \item \textbf{Accommodate different ontologies}: There are many equally valid categorizations for the same set of entities (e.g. there are multiple valid categorizations of bills by topic) that ought to be available. Some users might also want to define their own categorizations (e.g. considering all donors from the same address as a single entity).
    \item \textbf{Integration of user preferences and knowledge}: This includes adapting what the platform shows users based on their preferences, as well as users inputting their own knowledge into the system to run new analyses.

\end{enumerate}

\subsubsection{Usability and Analytics}

\begin{enumerate}
\setcounter{enumi}{5}

    \item \textbf{Adherence to UX best practices}: Though a rather basic requirement, we found that few platforms follow these best practices that reduce cognitive overhead and make the system more usable.
    \item \textbf{Allow follow up questions and flexible analyses paths}: Especially in the interconnected world of political information, data analyses are rarely a single-question endeavour; rather, they are explorations that follow non-linear paths.
    \item \textbf{Analysis of unstructured data}: Because much of the political data landscape involves unstructured data (e.g. social media, transcripts, bill text), systems should have a process in place to extract meaningful analysis out of these, be it automatically or by facilitating human intervention on these processes (e.g. building ontologies, training machine learning models).
    \item \textbf{Attend to different levels of user technical background}: Some users might be satisfied with simply obtaining data whereas others will want to ask more complex questions about the data that does not require them to manage the data itself. For each user group, the main use cases are:

    % \item Adherence to UX best practices. Though a rather basic requirement, we found that few platforms follow these best practices that reduce cognitive overhead and make the system more usable.
    % \item Allow follow up questions and flexible analyses paths. Especially in the interconnected world of political information, data analyses are rarely a single-question endeavour; rather, they are explorations that follow non-linear paths.
    % \item Analysis of unstructured data. Because much of the political data landscape involves unstructured data (e.g. social media, transcripts, bill text), systems should have a process in place to extract meaningful analysis out of these, be it automatically or by facilitating human intervention on these processes (e.g. building ontologies, training machine learning models).
    % \item Attend different levels of user technical background. Some users might be satisfied with simply obtaining data whereas others will want to ask more complex questions about the data that does not require them to manage the data itself. For each user group, the main use cases are:
    \begin{enumerate}
        \item \textit{Technically-savvy domain experts}: Allowing users to download the data to use it on other platforms.
        \item \textit{Domain experts without technical skills}: Aiding users with conducting analyses.
        \item \textit{Users without domain or technical expertise}: Identifying user questions and mapping these to analyses, as well as providing technical and domain context.

    \end{enumerate}

    \item \textbf{Export analysis}: To maximize the utility of the information, users should be able to share the information or use it in other platforms.

\end{enumerate}

\subsubsection{Transparency and Trust}

\begin{enumerate}
\setcounter{enumi}{10}
    \item \textbf{Transparent analysis and processing}: Documentation on analysis procedures, AI techniques, data scope, and data processing should be readily available. Building trustworthy and explainable AI is especially important  when handling sensitive data such as electoral finances and records.
    \item \textbf{Identification and communication of limitations}: This includes reporting limits on technical insight (e.g. uncertainty in analyses, strength of correlation) as well as domain significance (e.g. when looking at finances it is important to look at contributions and expenditures). This is especially pertinent to the system’s AI, since these are known to have significant issues with fairness and generalization outside their training data,
    \item \textbf{Allow verification of results by domain or technical experts}: Though we want to allow flexibility and lower barriers to entry, politics is a domain rife with misleading statements and outright fake information. Hence, there need to be balancing mechanisms to identify and highlight trustworthy content.

\end{enumerate}

\subsection{Use Cases following the System Requirements}

Consider a user interested in exploring trends in financial contributions to state congressional candidates by environmental interest groups in the state of Illinois. Given that the user knows a bit about the domain, they are interested in total contributions by environmental groups donated to each candidate committee, and different breakdowns (e.g. by time, by district, by political party). The user asks the platform these analytical questions, and the platform provides answers in graphs or text. The platform also identifies potential follow up analytical questions that might interest the user.

Consider a different user who has little knowledge about the domain, but wants to know more about their representatives in the state Congress and their record regarding immigration. The platform deduces the user’s representatives based on their address. Given the users’ limited domain knowledge, the platform also gives an overview of different roles (e.g. difference between senators and representatives) and definitions (e.g. what is a PAC). The platform then provides an overview of the representatives’ finances, voting records, and statements that relate to immigration.

\section{Conclusions and Future Work}

In this work we have identified the main sociotechnical requirements political information systems based on user interviews with domain stakeholders. These principles can be used to develop platforms for enabling users with interest in the domain to answer questions about politics. Development of said platform will be explored in future work.

A significant group of users that we did not interview were users who had no interest in politics. We excluded this group to be able to better focus on those who we do not need to persuade to seek political information, and due to inherent problems in trying to interview users who are not interested in politics about politics. Though outside the scope of this work, identifying requirements for these users is an imperative line of work that we also leave for future work.

\bibliography{references}

\end{document}